\documentclass[lettersize,journal]{IEEEtran}
\usepackage{amsmath,amsfonts}
\usepackage{algorithmic}
\usepackage{algorithm}
\usepackage{array}
\usepackage[caption=false,font=normalsize,labelfont=sf,textfont=sf]{subfig}
\usepackage{textcomp}
\usepackage{stfloats}
\usepackage{url}
\usepackage{verbatim}
\usepackage{graphicx}
\usepackage{cite}
\usepackage{booktabs}
\usepackage{makecell}
\usepackage{pifont}
\usepackage{xcolor,colortbl}
\usepackage{bbm}

\begin{document}

\title{Improve Cross-domain Mixed Sampling with Guidance Training for Adaptive Segmentation}

\author{Wenlve Zhou, Zhiheng Zhou *, Tianlei Wang, Delu Zeng
\thanks{This work is supported by the National Key Research and Development Program of China(2022YFF0607001), Guangdong Basic and Applied Basic Research Foundation (2023A1515010993), Guangdong Provincial Key Laboratory of Human Digital Twin (2022B1212010004), Guangzhou City Science and Technology Research Projects (2023B01J0011), Jiangmen Science and Technology Research Projects (2021080200070009151), Shaoguan Science and Technology Research Project(230316116276286), Foshan Science and Technology Research Project(2220001018608). (* Corresponding author: Zhiheng Zhou.)

Wenlve Zhou, Zhiheng Zhou, Delu Zeng are with School of Electronic and Information Engineering, South China University of Technology, Guangzhou 510641, Guangdong, China, and also with Key Laboratory of Big Data and Intelligent Robot, Ministry of Education, South China University of Technology, Guangzhou 510641, China (email: wenlvezhou@163.com; zhouzh@scut.edu.cn; dlzeng@scut.edu.cn).

Tianlei Wang is with the Department of Intelligent Manufacturing, Wuyi University, Jiangmen 529020, China (email: tianlei.wang@aliyun.com).
}}

\markboth{Journal of \LaTeX\ Class Files,~Vol.~14, No.~8, August~2021}%
{Shell \MakeLowercase{\textit{et al.}}: A Sample Article Using IEEEtran.cls for IEEE Journals}

\IEEEpubid{0000--0000/00\$00.00~\copyright~2021 IEEE}

\maketitle

\begin{abstract}
Unsupervised Domain Adaptation (UDA) endeavors to adjust models trained on a source domain to perform well on a target domain without requiring additional annotations. In the context of domain adaptive semantic segmentation, which tackles UDA for dense prediction, the goal is to circumvent the need for costly pixel-level annotations. Typically, various prevailing methods baseline rely on constructing intermediate domains via cross-domain mixed sampling techniques to mitigate the performance decline caused by domain gaps. However, such approaches generate synthetic data that diverge from real-world distributions, potentially leading the model astray from the true target distribution. To address this challenge, we propose a novel auxiliary task called Guidance Training. This task facilitates the effective utilization of cross-domain mixed sampling techniques while mitigating distribution shifts from the real world. Specifically, Guidance Training guides the model to extract and reconstruct the target-domain feature distribution from mixed data, followed by decoding the reconstructed target-domain features to make pseudo-label predictions. Importantly, integrating Guidance Training incurs minimal training overhead and imposes no additional inference burden. We demonstrate the efficacy of our approach by integrating it with existing methods, consistently improving performance. The implementation will be available at https://github.com/Wenlve-Zhou/Guidance-Training.
\end{abstract}


\section{Introduction}
\IEEEPARstart{I}{n} recent years, the advancement of deep learning has revolutionized various computer vision tasks \cite{ref1, ref2, ref3}. However, the inherent challenge of neural networks in handling out-of-distribution (OOD) scenarios, where a significant shift between training and testing distributions leads to drastic performance degradation, remains a critical concern. Conventional solutions often resort to re-collecting extensive labeled data aligned with the testing domain and retraining or fine-tuning neural networks accordingly. Yet, building large-scale annotated datasets is a time-consuming endeavor, particularly in semantic segmentation. Fortunately, the burgeoning interest in Unsupervised Domain Adaptation (UDA) has yielded pivotal progress \cite{ref4}, addressing the limitations of deep learning sensitive to domain gaps. By leveraging annotated source domain data and unlabeled target domain data, UDA empowers neural networks to achieve exceptional performance in the target domain without necessitating annotations.
\begin{figure}[!t]
	\setlength{\abovecaptionskip}{0.cm}
	\setlength{\belowcaptionskip}{-0.cm}
	\centering
	\includegraphics[width=3.5in]{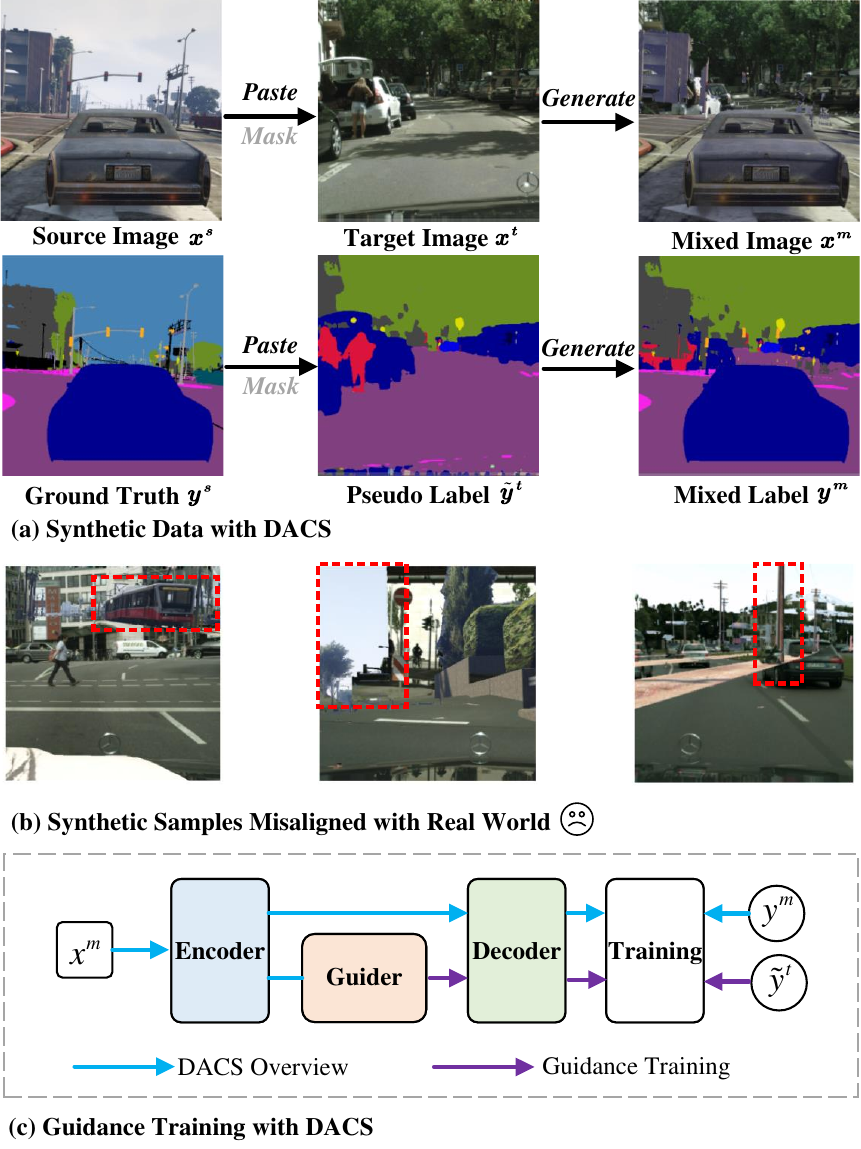}
	\caption{(a) The ground truth is randomly sampled to generate a binary mask, which serves as the basis for generating a hybrid image and its corresponding annotations through a copy-paste mechanism. (b) DACS-generated samples stray from physical norms, highlighting contextually deficient areas with red delineations. (c) Building upon DACS, we introduce the Guider module to guide the model in predicting the pseudo-labels of the original target image.}
	\label{fig_1}
\end{figure}
\IEEEpubidadjcol

In the realm of unsupervised domain adaptation for segmentation tasks \cite{ref5}, numerous notable works have emerged, progressively eliminating the gap toward segmentation performance achieved through supervised learning. Notably, these approaches adopt data mixing \cite{ref6,ref7,ref8,ref9} as a fundamental baseline, with ``Domain Adaptation via Cross-domain mixed Sampling'' (DACS)\cite{ref9} representing one of the most widely embraced methodologies. Specifically, DACS, as an exemplar of cross-domain mixing, involves random sampling of masks from the ground truth of the source domain to overlay pixels from the corresponding source domain image onto an image from the target domain, creating mixed images. Simultaneously, labels are mixed based on these masks, substituting the missing annotations in the target domain with predicted pseudo-labels, as illustrated in Figure 1(a). DACS constructs an intermediate domain by blending source and target domain images, effectively alleviating performance degradation caused by significant domain gaps. Owing to its convenience and efficiency, DACS has become a widely adopted baseline, applied across semantic segmentation\cite{ref10, ref11,ref12}, panoptic segmentation\cite{ref13}, and even in scenarios involving event cameras\cite{ref14}. Hence, in this paper, DACS is primarily analyzed and explored as a representative example of cross-domain mixed sampling.

Despite its considerable advantages, the random ``copy-paste'' strategy employed by DACS disrupts the fundamental characteristic of semantic segmentation tasks—the coherence of image context. As depicted in Figure 1(b), three images synthesized by DACS are randomly sampled, and areas lacking contextual coherence are highlighted within red boxes. In the left image, a train is depicted traversing through mid-air; in the middle image, the sky region within the red box abruptly appears on the road; in the right image, a pole is inserted into a car. These scenes, defying common sense, disrupt the model's learning of rules in the physical world, reducing its robustness and potentially posing safety risks, particularly in scenarios like autonomous driving.

Given the flexibility of DACS-based methodologies, our endeavor aims to keep the network from biasing the real-world distribution solely by introducing a novel auxiliary task, without altering the existing techniques. During the training of the hybrid image generated by DACS, we introduce a module called ``Guider'' between the encoder and decoder of the semantic segmentation model. This Guider is designed to guide the model in predicting pseudo-labels for the original target image based on the features extracted from the hybrid image. \textit{Given the similarity of this optimization objective to avoiding misled human guidance, we refer to this auxiliary task as ``Guidance Training'' in this paper, as depicted in Figure 1(c).} Our design motivation is to utilize Guidance Training as a constraint during training with mixed data, thereby overcoming domain gaps and ensuring alignment with real-world distributions. \textbf{Our approach offers two key advantages. Firstly, efficient. The introduction of Guider and the Guidance Training has minimal impact on GPU memory and training time. Secondly, pluggable. During the inference stage, Guider is removed, resulting in no additional computational overhead.}

Our contributions are summarized as follows: 	

(1) We propose Guidance Training to improve cross-domain mixed sampling to prevent model prediction from biasing real-world distributions.

(2) Our approach adds minimal GPU memory and training time overhead, with no additional computational burden during inference.

(3) Guidance Training is simple to implement and can be seamlessly integrated with various domain adaptive segmentation methods, consistently improving performance.

The subsequent sections of this paper are organized as follows: Section II presents a comprehensive literature review. In Section III, we introduce our proposed methodologies. Section IV provides detailed information regarding the experimental setup and presents the evaluation results. Finally, Section V concludes the paper and outlines potential directions for future research.

\begin{figure*}[!t]
	\setlength{\abovecaptionskip}{0cm}  
	\setlength{\belowcaptionskip}{-0.2cm} 
	\centering
	\includegraphics[width=6.8in]{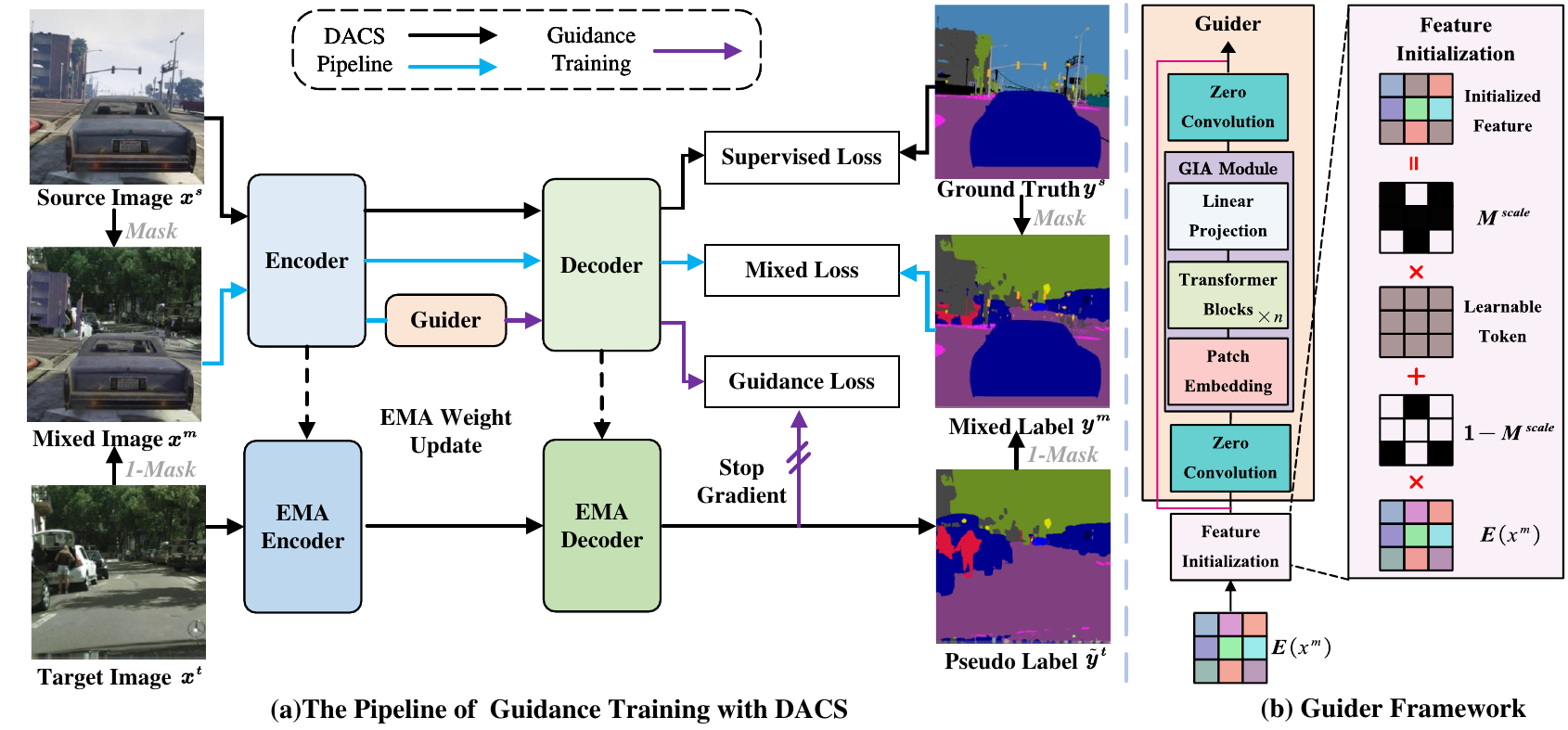}
	\caption{Guidance Training implemented with DACS \cite{ref9}. (a) Building upon the DACS pipeline, we incorporate the Guider between the encoder and decoder, steering model training via Guidance Loss, thereby constituting Guidance Training. (b) The primary role of the Guider is to aid the model in predicting pseudo-labels based on the hybrid features $E(x^{m})$, prompting the model to transition from the hybrid feature distribution to decouple it from the real-world feature distribution. The Guider is exclusively employed during training and therefore does not amplify model inference overhead.}
	\label{Fig2}
\end{figure*}

\section{Related Work}
In this section, we present a brief survey on the most related fields of our works in four aspects: semantic segmentation, unsupervised domain adaption, mixed sampling and masked image modeling.

\subsection{Semantic Segmentation}
Semantic segmentation, a foundational task in computer vision, has made remarkable strides, thanks to advancements in deep learning methodologies. Initially, Fully Convolutional Networks (FCNs) \cite{ref18} represented a significant breakthrough by enabling a granular understanding of semantic information at the pixel level. Nonetheless, persistent challenges revolve around accurately delineating small objects and navigating through intricate scenes. Moving beyond FCNs, the evolution of semantic segmentation design embraced the encoder-decoder structure \cite{ref19,ref20, ref21, ref22}, leading to substantial performance improvement. These improved techniques include skip-connections \cite{ref19}, dilated convolutions \cite{ref21}, etc. Previous methods relied on Convolution Neural Networks (CNNs) \cite{ref23}, while recent advancements introduced the Transformer architecture \cite{ref24}, renowned for its robust contextual modeling prowess, significantly enhancing performance in semantic segmentation tasks \cite{ref25}. Utilizing the Transformer architecture for high-resolution image tasks poses challenges due to the substantial computational overhead demanded by its core module, the self-attention mechanism \cite{ref24}. To address this issue, Xie \textit{et al.} \cite{ref26} implemented separate downsampling techniques for the key and value within the self-attention mechanism, effectively mitigating the burden. Inspired by ConvNeXt \cite{ref27}, SegNext \cite{ref28} devises a CNN-like backbone integrating linear self-attention, demonstrating robust performance across various semantic segmentation benchmarks. 

\subsection{Unsupervised Domain Adaption}
Unsupervised domain adaptation has emerged as a significant focus within the realm of computer vision, with the primary goal of addressing the challenge posed by domain shift. UDA methods aim to utilize labeled data from a source domain to enhance the performance of a model operating in a target domain where labeled data is scarce or unavailable. A variety of methodologies have been proposed to tackle adaptive classification, encompassing domain adversarial learning \cite{ref29,ref30}, self-training \cite{ref31}, and consistency regularization \cite{ref32}. These approaches strive to align the distributions of features between the source and target domains, thereby facilitating the transfer of knowledge. Given the demands of segmentation tasks in processing high-resolution images and contextual correlations, methodologies tailored for classification tasks often lack direct applicability to unsupervised adaptive segmentation tasks. DAFormer \cite{ref10} introduces several techniques such as rare class sampling and feature space regularization, effectively enhancing performance. Contrasting with prior approaches primarily focused on reducing memory usage through image resizing, HRDA \cite{ref11} introduces a multi-resolution training method. This approach aims to capture distant contextual dependencies, acknowledging that loss of details may potentially impair model performance. Building upon HRDA \cite{ref11}, PiPa \cite{ref33} introduces a comparative learning approach that compactly represents intra-class features while emphasizing discriminative inter-class features, effectively capturing contextual information within the model.

\subsection{Mixed Sampling}
Mixed sampling, a simple yet effective method of data augmentation, finds widespread application in Computer Vision (CV) \cite{ref6, ref7} and Natural Language Processing (NLP) \cite{ref47,ref48} across various learning paradigms, including supervised, semi-supervised, and unsupervised learning. In computer vision, the common approach involves creating a new mixed image by blending pixels from two images using different techniques. One classical method is mixup \cite{ref7}, where the two images are weighted with their corresponding labels, determined by sampling mixing factors from the beta distribution. Manifold Mixup \cite{ref49} extends this technique from mixing input data to interpolating neural network middle layer representations. In the segmentation task, the mixed data generation approach with pixel weighting yields only limited improvement in model performance. The augmented output of ClassMix \cite{ref40} consists of a blend of inputs, wherein half of the semantic classes from one image are overlaid onto the other. DACS \cite{ref9} incorporates the ClassMix \cite{ref40} into the adaptive segmentation task, resulting in a notable improvement in the performance of the UDA when combined with a straightforward student-teacher structure.

\subsection{Masked Image Modeling}
Masked modeling is a popular method for enhancing contextual relevance. Predicting reserved tokens in masked input sequences has proven to be a robust self-supervised pretraining task in NLP \cite{ref34,ref35}. Recently, this concept has been successfully transposed into self-supervised pretraining in computer vision, referred to as masked image modeling. Based on its training objectives, MIM can be categorized into Low-level and High-level reconstruction tasks. MAE \cite{ref17} and MaskFeat \cite{ref36} introduce a lightweight decoder to compel the model to predict masked original pixel values and HOG features respectively. Drawing inspiration from 3D point cloud processing, ConvNextV2 \cite{ref37} incorporates sparse convolution techniques to execute the MAE task within CNNs. In contrast, high-level tasks commonly involve auxiliary networks to create target features, guiding the backbone network during reconstruction training. BEIT \cite{ref15} achieves effective visual representation by training dVAE \cite{ref38} and follows Bert's \cite{ref34} approach to backbone training. Meanwhile, i-JEPA \cite{ref39} guides intermediate layer feature prediction through the introduction of a teacher network.

This section illustrates the extensive efforts directed towards enriching model contextual information to enhance segmentation performance in both supervised and UDA semantic segmentation tasks. While DACS constructs intermediate domains, thus bolstering network performance in the target domain, our earlier discussion in Section I highlights the contextual structure degradation in the images synthesized by DACS. To mitigate this issue, we introduce Guidance Training to align the model's predictive distribution with the physical world, with the aim of enhancing domain adaptive segmentation performance and improving model robustness.

\section{Methods}
\subsection{Preliminary: DACS}
In unsupervised domain adaptive segmentation, a neural network undergoes training using labeled data from the source domain $D_{s}=\left\{\left(x_{k}^{s}, y_{k}^{s}\right)\right\}_{k=1}^{n_{s}}$ to attain optimal performance on the target domain $D_{t}=\left\{\left(x_{k}^{t}\right)\right\}_{k=1}^{n_{t}}$ without access to the target labels. Within semantic segmentation networks, $F(\cdot )$ typically comprises an Encoder $E(\cdot )$ and a Decoder $D(\cdot )$, represented as $F(\cdot )=E(\cdot )\circ D(\cdot )$. Simply training the network using categorical cross-entropy (CE) loss on the source domain can be formulated as:
\begin{eqnarray}
	\mathcal{L}_{sup}=-\sum_{i=1}^{H} \sum_{j=1}^{W} \sum_{c=1}^{C} y^{s}_{i j c} \log F(x^s)_{i j c}
\end{eqnarray}
where $H$ and $W$ represent the image's height and width respectively, while $C$ signifies the number of categories in the UDA task.

However, the model struggles with generalization to the target domain, primarily due to feature bias in test scenarios. To bridge the domain gap, DACS \cite{ref9} introduces intermediate domains by employing mixing strategies \cite{ref40} that guide model training with a mixing loss formulation:
\begin{eqnarray}
	\mathcal{L}_{mix}=-\sum_{i=1}^{H} \sum_{j=1}^{W} w_{i j}  \sum_{c=1}^{C} y^{m}_{i j c} \log F(x^m)_{i j c}
\end{eqnarray}
where $w_{i j}$ represents the dynamic trade-off (which will be describe later). From Equation (2), the training methodology of DACS aligns with the supervised part, differing primarily in the creation of the mixed image $x^m$ and its corresponding label $y^m$. Following ClassMix \cite{ref40}, given an annotation $y^{s}$, half of the different classes present in $y^{s}$ are randomly selected, forming the set of sampled classes $S$. Utilizing this set, a binary mask $M$ is constructed as outlined below:
\begin{eqnarray}
	M(i, j)=\left\{\begin{array}{l}
	1, \text { if } {y}^{s}(i, j) \in S \\
	0, \text { otherwise }
\end{array}\right.
\end{eqnarray}
Upon the generated Mask, a mixed image with its label is defined:
\begin{eqnarray}
\begin{aligned}
	x^{m} & =M \odot x^{s}+(1-M) \odot x^{t} \\
	y^{m} & =M \odot y^{s}+(1-M) \odot \tilde{y}^{t}
\end{aligned}
\end{eqnarray}
where $\odot$ denotes element-wise multiplication and $\tilde{y}_{t}$ represents the pseudo label of $x_{t}$. To better transfer the knowledge from the source to the target domain, DACS use a teacher network $\overline{F}(\cdot )=\overline{E}(\cdot )\circ \overline{D}(\cdot )$ to produce pseudo-labels for the target domain data. 
\begin{eqnarray}
\tilde{y}^{t}_{i j}=\underset{c}{argmax}\text{ } \overline{F}(x^t)_{i j c}
\end{eqnarray}
Commonly, the teacher model $\overline{F}$ are set as the exponentially moving average of the weights of $F$ after each training step $T$ \cite{ref41} to increase the stability of the predictions.
\begin{eqnarray}
\overline{F}\gets \alpha \overline{F} + (1-\alpha)F_{T}
\end{eqnarray}

Given the validity of pseudo-label in mixed images, the introduced weights $w_{i j}$ in Equation (2) can dynamically adjust the training intensity.
\begin{eqnarray}
\begin{aligned}
	w_{i j} & = 1 \cdot M + q \cdot (1-M)
\end{aligned}
\end{eqnarray}
\begin{eqnarray}
	\begin{aligned}
		q = \frac{\sum_{i=1}^{H}\sum_{j=1}^{W}\left[\mathbbm{1}_{[\tau<\max _{c} \overline{F}\left(x^{t}\right)_{ijc}]}\right]}{H \cdot W}
	\end{aligned}
\end{eqnarray}
where $\mathbbm{1}_{[\cdot]}$ is the indicator function. Here, $\tau$ represents a predefined threshold used to filter reliable predictions. The overall objective of DACS is formulated as:
\begin{eqnarray}
	\mathcal{L}_{dacs} = \mathcal{L}_{sup} + \mathcal{L}_{mix}
\end{eqnarray}

\subsection{Guidance Training}
In computer vision tasks, models need to assimilate image contextual information for effective semantic parsing, a crucial aspect in tasks like semantic segmentation. For instance, when a person is seated on a bicycle, they're identified as a rider, but when on a sidewalk, they're recognized as a pedestrian. DACS \cite{ref9}, by randomly mixing images, compromises the contextual information within the image structure, making it challenging for the model to grasp real-world distributions, as illustrated in Fig. 1(b). Therefore, our proposal aims to guide the model to understand the contextual relationships within the original target image, leveraging target domain information in the blended images. It ensures that the model is aligned with the real-world distribution while overcoming the domain gap. This approach provides additional cues for robustly recognizing classes that share similar local appearances.

Learning contextual information based on local semantics by introducing MAE \cite{ref17} as an auxiliary task seems intuitive. The input image $x$ is reshaped into non-overlapping patches $p=\left\{p_{i}\right\}_{i=1}^{N_l}$. MAE constructs a random mask $M^{mae} \in\{0,1\}^{N_l}$ to indicate the masked patches, where $M^{mae}_{i}=1$ corresponds to the patches that are masked. Only the visible patches $p^{v}=\left\{p_{i} \mid M^{mae}_{i}=0\right\}_{i=1}^{N_v}$ are fed into the encoder and are mapped to potential features, after which a lightweight decoder $D_{mae}$ is introduced to reconstruct the pixel values of the invisible patches. 
\begin{eqnarray}
	\mathcal{L}_{mae}=-\sum_{i=1}^{N_{l}} M_{i}^{mae} \cdot \left ( D_{mae}(E(p^{v})) - p \right )^{2}
\end{eqnarray}
However, the introduction of MAE faces three major challenges: \textbf{1)} MAE, functioning as an auxiliary task, can solely train the encoder $E(\cdot )$ while leaving the decoder $D(\cdot )$ untrained; \textbf{2)} Apart from $x^{s}$ and $x^{m}$, additionally encoding $p^{v}$ in the encoder results in a significant computational overhead due to the extensive parameter count in the encoder. \textbf{3)} MAE cannot be directly applied to convolutional neural networks.

To tackle the aforementioned challenges, we propose two improvement strategies: \textbf{1)} Changing the optimization objective from predicting pixel values to predicting image labels, thereby involving the decoder in the context prediction training; \textbf{2)} To avoid additional coding, our plan involves leveraging the latent features of $x^m$. Within these latent features, we conceptualize the features from the source domain as a specialized mask for the features within the target domain. This approach allows us to reconstruct the features specific to the target domain image by introducing a Guider $G(\cdot )$ (refer to Section III.C for detailed information). Subsequently, these reconstructed features are integrated into the decoder. The guidance loss $\mathcal{L}_{gt}$ is then represented as:
\begin{eqnarray}
	\mathcal{L}_{gt}=-\beta\sum_{i=1}^{H} \sum_{j=1}^{W} q_{i j}  \sum_{c=1}^{C} \tilde{y}^{t}_{i j c} \log D(G(E(x^m),M))_{i j c}
\end{eqnarray}

Operations such as MAE use a fixed ratio for masking, whereas in Guidance Training, the masking ratio (i.e., the proportion of the source feature occupying the mixture of features) varies. When the masking ratio is higher, the uncertainty increases, making it more challenging to predict the pseudo-labels of the original target image. This heightened difficulty may lead the model to exhibit a bias towards random prediction outcomes, thereby compromising overall model performance. To accommodate this variability, we propose introducing an adaptive factor $\beta$ to estimate uncertainty. This adaptive factor aims to decrease when the masking ratio is high, allowing for dynamic adjustment of training intensity.
\begin{eqnarray}
	\begin{aligned}
		\beta = 1-exp(-d \cdot r)
	\end{aligned}
\end{eqnarray}
\begin{eqnarray}
	\begin{aligned}
		r = \frac{\sum_{i=1}^{H}\sum_{j=1}^{W}\left[\mathbbm{1}_{[M_{i j}=1]} \right]}{H \cdot W}
	\end{aligned}
\end{eqnarray}
The percentage of source component in the mask is tallied to calculate the ratio $r$. Subsequently, an exponential function is utilized to transform $r$ into a range between 0 and 1. In this paper, a constant factor $d$ is introduced, maintaining a fixed value of 5. Finally, the overall objective is formulated as:
\begin{eqnarray}
	\mathcal{L}_{overall} = \mathcal{L}_{dacs} + \lambda_{gt} \mathcal{L}_{gt}
\end{eqnarray}
Here, $\lambda_{gt}$ is the trade-off in Guidance Training.

\subsection{Guider}
The goal of Guider is to transform the hybrid feature $f^{m} = E(x^{m})$ into a pseudo target feature $\tilde{f}^{t} = G(f^{m},M)$, enabling subsequent decoder $D(\cdot )$ to predict the corresponding pseudo labels $\tilde{y}^{t}$. To enable $D(\tilde{f}^{t})$ to accurately predict pseudo-labels, the simplest approach involves introducing an alignment loss, inspired by DSN \cite{ref42}, to minimize the discrepancy between $\tilde{f}^{t}$ and $f^{t} = E(x^{t})$. However, this method poses a challenge as it can compromise the representation of the hybrid feature $f^{m}$, failing to effectively leverage the intermediate domain constructed by DACS \cite{ref9} and address the domain gap. Therefore, we opt for an alternative strategy—generating the pseudo target features with a more loosely constrained distribution alignment objective. This allows the model to impose constraints on the feature space through contextual predictions while effectively utilizing the intermediate domain. The goal is to avoid overfitting on mixed data, preventing biases in learning the distributions of real-world physical rules.

To avoids introducing significant extra training overhead, Guider maintain a efficient architectural design. Common distributional alignment losses \cite{ref43, ref44} might struggle to achieve effective distributional alignment or be prone to converging to trivial solutions. Consequently, we modify the role of Guider, shifting from learning the entirety of $\tilde{f}^{t}$ to focusing on learning the offset of the initialized pseudo target feature $f^{ini}$ (red line in Figure 2b). Following training, the initialized features can be regarded as embryonic form of the target features. This adjustment alleviates the pressure on Guider's representation.
\begin{eqnarray}
	\tilde{f}^{t}=G(f^{ini})+f^{ini}
\end{eqnarray}
During the feature initialization phase, inspired by MAE \cite{ref17}, we treat the features from the source image in the mix features as a unique type of mask for the target features. Initially, we eliminate $p^{s}=\left\{f^{m}_{ij} \mid M^{scale}_{ij}=1\right\}_{i=1}^{N_s}$ and insert learnable tokens $t$ into the corresponding positions. Due to the size disparity between the mask $M$ and the feature map $f^{m}$, we downsample the mask $M$, resulting in the creation of $M^{scale}$ (details in ``Feature Initialization'' of Figure 2b).
\begin{eqnarray}
f^{ini} = M^{scale} \odot t +(1-M^{scale}) \odot f^{m}
\end{eqnarray}

After the initialization features are constructed, the Global Information Aggregation Module (GIA) $A(\cdot)$ is introduced. This module comprises patch embedding \cite{ref45}, numbers of Transformer blocks \cite{ref45}, and a linear projection layer. The patch embedding primarily serves to reduce spatial and channel dimensions, promoting model convergence. Transformer blocks are employed to integrate global context information for offset learning. Finally, the linear projection layer scales back the spatial and channel dimensions of the original feature map. After patch embedding, we added cosine positional encoding \cite{ref24} to each patch. To stabilize the training process, two zero-initialized 1*1 convolution, $Z_1(\cdot)$ and $Z_2(\cdot)$, is introduced in the input and output sections, following the ControlNet \cite{ref46}. The details are shown in Fig. 2b.
\begin{eqnarray}
	G(f^{ini})= Z_{2}(A(Z_{1}(f^{ini})))
\end{eqnarray}

We engineered Guider’s architecture to be composed of a minimal number of components, and Guider exhibit shallow properties, thereby resulting in a negligible increase in training time and GPU memory usage when introducing Guider and Guidance Training (details in Table II). Crucially, it exclusively engages in model training and can be seamlessly excluded during inference, without imposing any added burden on the inference process of the model.

\section{Experiments}
In this section, we primarily focus on experimenting with two popular benchmarks: GTA$\rightarrow$Cityscape and Synthia$\rightarrow$Cityscape. In Subsection IV.A, we introduce the dataset along with implementation details. Subsection IV.B delves into the analysis of  Guidance Training combined with the popular UDA methods, including improvements, training time, and GPU memory usage. Subsection IV.C involves a comparison with State-of-the-Art (SoTA) approaches. Subsection IV.D explores the performance impact under various parameter settings through ablation studies. Finally, in Subsection IV.E, we present a qualitative analysis of our method.

\subsection{Experiment Setups}
\textbf{Dataset.} The GTA \cite{ref50} dataset comprises 24,966 synthetic images featuring pixel-level semantic annotations. These images are generated within the open-world environment of ``Grand Theft Auto V'', all captured from the perspective of a vehicle navigating the streets of American-style virtual cities. This dataset encompasses 19 semantic classes that align with those found in the Cityscapes dataset.

SYNTHIA \cite{ref51} constitutes a synthetic urban scene dataset. We opt for its subset known as ``SYNTHIA-RAND-CITYSCAPES'', which shares 16 common semantic annotations with Cityscapes. Specifically, we utilize a total of 9,400 images, each with a resolution of 1280×760, sourced from the SYNTHIA dataset.

Cityscapes \cite{ref52} is a dataset featuring real urban scenes captured across 50 cities in Germany and neighboring regions. The dataset includes meticulously annotated images, comprising 2,975 training images, 500 validation images, and 1,525 test images, all at a resolution of 2048×1024 pixels. Each pixel within these images is classified into one of 19 distinct categories.

\textbf{Implementation Details.} For network architecture, we utilize the DeepLab-V2 \cite{ref21} with ResNet101 \cite{ref53} as the backbone. As a standard practice, we adhere to the self-training approach outlined in DAFormer \cite{ref10} along with its associated training parameters. Specifically, we employ AdamW \cite{ref54} optimization with a learning rate of $6 \times 10^{-5}$ for the encoder and $6 \times 10^{-4}$ for the decoder, utilizing a batch size of 2. Additionally, we incorporate linear learning rate warmup, and maintain a momentum parameter $\alpha$ of 0.999. The predefined threshold $\tau$ is set to 0.968.

For Guidance Training, we chose a trade-off $\lambda_{gt}$ value of 1.0. For uncertainty estimation, we set $d$ to 5. Regarding the design of the Guider, the patch size in the patch embedding is set to 4, the number of transformer blocks is set to 2, and the embedding dimension is set to 512. The learning rate of $6 \times 10^{-5}$ is employed with the Guider.  All the experiments are conducted with 1x NVIDIA GTX 3090 with 24G RAM and the PyTorch framework is implemented to perform our experiments.

\subsection{Various UDA Methods with Guidance Training}
\begin{table}[t]
	\setlength{\abovecaptionskip}{0cm}  
	\setlength{\belowcaptionskip}{-0.2cm} 
	\centering
	\caption{Various UDA Methods with Guidance Training for Performance Improvement on GTA$\rightarrow$Cityscape. * Denotes the Results of Experiments Conducted by Us.}
	\renewcommand{\arraystretch}{1.0}
	\setlength{\tabcolsep}{2.0mm}{
		\begin{tabular}{cccc}
			\toprule
			UDA Method & w/o Guidance & w/ Guidance & Improvement \\
			\midrule
			DACS$^{*}$ \cite{ref9} & 53.2 & 55.9 & \textcolor{teal}{+2.7} \\
			DAFormer \cite{ref10} & 56.0 & 58.5 & \textcolor{teal}{+2.5} \\
			HRDA \cite{ref11} & 63.0 & 64.4 & \textcolor{teal}{+1.4} \\
			MIC \cite{ref55} & 64.2 & 67.0 &  \textcolor{teal}{+2.8} \\
			\bottomrule
		\end{tabular}%
	}
	\label{tab1}%
\end{table}%

\definecolor{light-gray}{gray}{0.92}
\begin{table}[t]
	\setlength{\abovecaptionskip}{0cm}  
	\setlength{\belowcaptionskip}{-0.2cm} 
	\centering
	\caption{Various UDA Methods with Guidance Training for Computational Overhead on RTX3090.}
	\renewcommand{\arraystretch}{1.1}
	\setlength{\tabcolsep}{1.0mm}{
		\begin{tabular}{ccccc}
			\toprule
			&\multicolumn{2}{c}{Training}&\multicolumn{2}{c}{Inference} \\
			&Training Time & GPU Memory & Throughput & GPU Memory \\
			\midrule
			DACS \cite{ref9} & 7.25h & 20.5 GB & 12.1 img/s & 20.51 GB  \\
			\rowcolor{light-gray}
			w/ Guidance & 7.28h & 20.5 GB & 12.1 img/s & 20.51 GB\\
			\midrule
			DAFormer \cite{ref10} & 8.12h & 22.0 GB & 12.1 img/s  & 20.51 GB\\
			\rowcolor{light-gray}
			w/ Guidance &8.25h & 22.0 GB & 12.1 img/s & 20.51 GB\\
			\midrule
			HRDA \cite{ref11} & 25.67h & 22.16 GB & 1.2 img/s & 21.26 GB\\
			\rowcolor{light-gray}
			w/ Guidance & 25.72h & 22.26 GB  & 1.2 img/s & 21.26 GB \\
			\midrule
			MIC \cite{ref55} & 31.50h & 22.20 GB & 1.2 img/s  & 21.26 GB \\
			\rowcolor{light-gray}
			w/ Guidance & 31.90h & 22.44 GB & 1.2 img/s & 21.26 GB \\
			\bottomrule
		\end{tabular}%
	}
	\label{tab2}%
\end{table}%
Guidance Training is a simple and effective method that can be seamlessly integrated with various approaches. In this subsection, we examine recent popular domain adaptive segmentation methods \cite{ref9,ref10,ref11,ref55} that utilize cross-domain data mixing techniques as the baseline. We integrate Guidance Training with these methods to explore the performance enhancement it offers and analyze associated computational overhead.

Table I presents the results of experiments conducted on the GTA$\rightarrow$Cityscape benchmark using UDA methods without Guidance (w/o Guidance) and combined with Guidance Training (w/ Guidance). The experiments demonstrated that combining Guidance with popular UDA methods consistently resulted in performance gains, with an average increase of +2.3. These findings indicate that Guidance Training can effectively enhance cross-domain mixing-based methods.

In addition to evaluating performance improvements, we carefully examine the impact of introducing Guidance Training on the original UDA method in terms of computational overhead, analyzing both the training and inference phases separately. Thanks to the efficient design of Guider, we observe minimal increase in training duration and GPU memory usage during the training period. Compared to the original UDA method, the introduction of Guidance Training results in an average increase of +0.15 hours (h) in training time and +0.09 GB in memory usage. However, there is no change in throughput compared to the original method, thanks to the plug-and-play feature of Guider.

\subsection{Comparisons with State-of-the-Arts UDA Methods}
\begin{table*}[t]
	\setlength{\abovecaptionskip}{0.cm}
	\setlength{\belowcaptionskip}{-0.cm}
	\centering
	\caption{Comparison with State-of-the-Art Methods for UDA. All Methods are based on DeepLab-V2 with ResNet-101 for a Fair Comparison. * Denotes the Results of Experiments Conducted by Us.}
	\renewcommand{\arraystretch}{1.0}
	\setlength{\tabcolsep}{0.8mm}{
		\begin{tabular}{c|ccccccccccccccccccc|c}
			\toprule
			Method &Road &S.walk &Build. &Wall &Fence &Pole &Tr.Light &Sign &Veget. &Terrain &Sky &Person &Rider &Car &Truck &Bus &Train &M.bike &Bike &mIOU\\
			\midrule
			\multicolumn{21}{c}{GTA$\rightarrow$Cityscape}\\
			\midrule
			IAST \cite{ref56} & 94.1& 58.8& 85.4& 39.7& 29.2& 25.1& 43.1& 34.2& 84.8& 34.6& 88.7& 62.7& 30.3& 87.6& 42.3& 50.3& 24.7& 35.2& 40.2& 52.2\\
			UPLR \cite{ref57} & 90.5& 38.7& 86.5& 41.1& 32.9& 40.5& 48.2& 42.1& 86.5& 36.8& 84.2& 64.5& 38.1& 87.2& 34.8& 50.4& 0.2& 41.8& 54.6& 52.6 \\
			DPL-dual \cite{ref58} & 92.8& 54.4& 86.2& 41.6& 32.7& 36.4& 49.0& 34.0& 85.8& 41.3& 86.0& 63.2& 34.2& 87.2& 39.3& 44.5& 18.7& 42.6& 43.1& 53.3\\
			SAC \cite{ref59}& 90.4& 53.9& 86.6& 42.4& 27.3& 45.1& 48.5& 42.7& 87.4& 40.1& 86.1& 67.5& 29.7& 88.5& 49.1& 54.6& 9.8& 26.6& 45.3& 53.8\\
			CTF \cite{ref60} & 92.5& 58.3& 86.5& 27.4& 28.8& 38.1& 46.7& 42.5& 85.4& 38.4& \textbf{91.8}& 66.4& 37.0& 87.8& 40.7& 52.4& 44.6& 41.7& 59.0& 56.1\\
			CorDA \cite{ref61} & 94.7& 63.1& 87.6& 30.7& 40.6& 40.2& 47.8& 51.6& 87.6& 47.0& 89.7& 66.7& 35.9& 90.2& 48.9& 57.5& 0.0& 39.8& 56.0& 56.6\\
			ProDA \cite{ref62}& 87.8& 56.0& 79.7& 46.3& 44.8& 45.6& 53.5& 53.5& 88.6& 45.2& 82.1& 70.7& 39.2& 88.8& 45.5& 59.4& 1.0& 48.9& 56.4& 57.5\\
			SePiCo \cite{ref63}& 95.2& 67.8& 88.7& 41.4& 38.4& 43.4& 55.5& \textbf{63.2}& 88.6& 46.4& 88.3& 73.1& 49.0& 91.4& 63.2& 60.4& 0.0& 45.2& 60.0& 61.0\\
			\midrule
			DACS \cite{ref9}& 89.9& 39.7& 87.9& 30.7& 39.5& 38.5& 46.4& 52.8& 88.0& 44.0& 88.8& 67.2& 35.8& 84.5& 45.7& 50.2& 0.0& 27.3& 34.0& 52.1\\
			\rowcolor{light-gray}
			w/ Guidance& 94.4 &62.8 &87.6 &35.7 &38.4 &38.0 &51.1 &58.4 &87.5 &45.7 &85.7 &68.9 &32.2 &90.0 &56.2 &62.4 &0.0 &21.9 &45.8 &55.9\\
			DFormer$^{*}$ \cite{ref10}& 95.7 & 70.2 & 87.8 &36.2 & 35.2 & 38.3& 50.2 & 53.8 &88.1 & 45.5 & 88.3 &69.6 &43.7 &87.7 &49.4 &50.2 &0.0 &24.9 &48.3 &56.0\\
			\rowcolor{light-gray}
			w/ Guidance& 95.4 &68.1 &88.3 &39.3 & 40.1 &38.9 &52.5 &54.1 &88.3 &45.6 &88.2 &70.6 &45.2 &87.1 &57.2 &53.6 &1.1 &40.9 &56.2 &58.5\\
			HRDA \cite{ref11}& 96.2& 73.1& 89.7& 43.2& 39.9& 47.5& 60.0& 60.0& 89.9& 47.1& 90.2& 75.9& 49.0& 91.8& 61.9& 59.3& 10.2& 47.0& 65.3& 63.0\\
			\rowcolor{light-gray}
			w/ Guidance& \textbf{96.5} &74.0 &90.2 &46.2 &43.6 &45.6 &60.4 &60.7 &90.0 &46.9 &89.5 &76.8 &48.0 &92.9 &\textbf{70.9} & 67.0 &4.4 &53.9 &65.8 &64.4\\
			MIC \cite{ref55}& \textbf{96.5}& 74.3 &90.4& 47.1& 42.8& \textbf{50.3}& 61.7& 62.3& \textbf{90.3}& \textbf{49.2}& 90.7& \textbf{77.8}& \textbf{53.2}& \textbf{93.0}& 66.2& 68.0& 6.8& 38.0& 60.6& 64.2\\
			\rowcolor{light-gray}
			w/ Guidance & 95.6 &\textbf{75.3} &\textbf{90.6} &\textbf{49.2} &\textbf{46.5} &49.5 &\textbf{61.8} &61.0 &\textbf{90.3} &48.6 &90.8 &76.9 &51.3 &92.9 &68.3 &\textbf{72.1} &\textbf{24.9} &\textbf{58.3} &\textbf{68.4} &\textbf{67.0}\\
			\midrule
			\multicolumn{21}{c}{Synthia$\rightarrow$Cityscape}\\
			\midrule
			DAF+IDA \cite{ref64}& 86.3& 41.6& 81.4& 18.3& 1.3& 25.1 &11.0 &10.5 &82.9 & – &85.6 &62.4 &27.1 &83.9 & – &55.9 & – &33.3 &41.2 &46.7\\
			UPLR \cite{ref57}& 79.4& 34.6& 83.5& 19.3& 2.8& 35.3& 32.1& 26.9& 78.8& – &79.6& 66.6& 30.3& 86.1& – &36.6 &– &19.5 &56.9 &48.0\\
			CTF \cite{ref60} &75.7 &30.0 &81.9 &11.5 &2.5 &35.3 &18.0 &32.7 &86.2 &– &90.1 &65.1 &33.2 &83.3 &– &36.5 &– &35.3 &54.3 &48.2\\
			IAST \cite{ref56}  &81.9 &41.5 &83.3 &17.7 &4.6 &32.3 &30.9 &28.8 &83.4 &– &85.0 &65.5 &30.8 &86.5 &– &38.2 &– &33.1 &52.7 &49.8\\
			SAC \cite{ref59} &89.3 &47.2 &85.5 &26.5 &1.3 &43.0 &45.5 &32.0 &87.1 &– &89.3 &63.6 &25.4 &86.9 &– &35.6 &– &30.4 &53.0 &52.6\\
			CorDA \cite{ref61} &\textbf{93.3} &61.6 &85.3 &19.6 &5.1 &37.8 &36.6 &42.8 &84.9 &– &90.4 &69.7 &41.8 &85.6 &– &38.4 &– &32.6 &53.9 &55.0\\
			ProDA \cite{ref62} &87.8 &45.7 &84.6 &37.1 &0.6 &44.0 &54.6 &37.0 &88.1 &– &84.4 &74.2 &24.3 &88.2 &– &51.1 &– &40.5 &45.6 &55.5\\
			SePiCo \cite{ref63} &77.0 &35.3 &85.1 &23.9 &\textbf{3.4} &38.0 &51.0 &55.1 &85.6 &– &80.5 &73.5 &46.3 &87.6 &– &\textbf{69.7} &– &50.9 &66.5 &58.1\\
			\midrule
			DACS \cite{ref9} &80.6 &25.1 &81.9 &21.5& 2.9 &37.2 &22.7 &24.0 &83.7 &– &\textbf{90.8} &67.6 &38.3 &82.9 &– &38.9 &– &28.5 &47.6 &48.3\\
			\rowcolor{light-gray}
			w/ Guidance & 88.8 & 50.4 &85.0 &14.1 &2.6 &36.9 &44.4 &44.2 &\textbf{88.2} &– &88.4 & 69.9 &45.0 &86.0 &– &48.9 &– &39.2 &61.6 &55.6\\
			DAFormer$^{*}$ \cite{ref10}& 72.5& 31.1 &84.6 &20.3 &2.5 &38.4 &44.3 &52.0 &84.0 &– &86.6 &69.6 &39.3 &86.1 &–  &54.6 &– &41.4 &59.0 &54.2\\
			\rowcolor{light-gray}
			w/ Guidance &78.9 &39.0 &84.9 &18.9 &2.5 &39.4 &47.3 &52.3 &79.8 &– &86.8 &69.9 &44.3 &85.5 &– &48.6 &– &40.5 &61.7 &55.0\\
			HRDA \cite{ref11} &85.8 &\textbf{47.3} &87.3 &27.3 &1.4 &50.5 &57.8 &61.0 &87.4 &– &89.1 &76.2 &48.5 &87.3 &– &49.3 &– &55.0 &68.2 &61.2\\
			\rowcolor{light-gray}
			w/ Guidance &84.5 &45.5 &87.0 &28.7 &2.7 &50.8 &61.0 &\textbf{62.6} &87.7 &– &88.5 &\textbf{78.8} &53.5 &89.1 &– &58.4 &– &\textbf{57.2} &68.8 &62.8\\
			MIC \cite{ref55} &84.7 &45.7 &\textbf{88.3} &29.9 &2.8 &\textbf{53.3} &61.0 &59.5 &86.9 &– &88.8 &78.2 &53.3 &89.4 &– &58.8 &– &56.0 &68.3 &62.8\\
			\rowcolor{light-gray}
			w/ Guidance  &84.7& 45.3& 88.0& \textbf{37.3}& 2.8& \textbf{53.3}& \textbf{62.5}& 61.8& 87.2& – &88.5 &78.3 &\textbf{53.7} &\textbf{90.4} &– &60.9 &– &56.1 &\textbf{71.4} &\textbf{63.8}\\
			\bottomrule
		\end{tabular}%
	}
	\label{tab3}%
\end{table*}%
In this subsection, we combine Guidance Training with UDA approaches \cite{ref9,ref10,ref11,ref55} based on cross-domain mix sampling and compare it with the state-of-the-art approach. Our experiments are conducted on two benchmarks, GTA$\rightarrow$Cityscape and Synthia$\rightarrow$Cityscape, respectively. The specific results are shown in Table III.

\textbf{GTA$\rightarrow$Cityscape.} We integrated Guidance Training with DACS \cite{ref9}, DAFormer \cite{ref10}, HRDA\cite{ref11}, and MIC \cite{ref55}, resulting in mIOU scores of 55.9, 58.5, 64.4, and 67.0, respectively. Compared to the baseline method, these enhancements yielded increases of +3.8, +2.5, +1.4, and +2.8, respectively. Alongside consistent performance enhancements, our analysis of experimental data revealed intriguing phenomena. Notably, the segmentation of motorcycle (M.bike) and bicycle (Bike), traditionally challenging due to their similarities, attained IOU scores of 58.3 and 68.4, respectively, when paired with MIC. This signifies significant improvements of +20.3 and +7.8 IOUs, respectively, compared to the baseline. Furthermore, our approach demonstrates competitive performance against the state-of-the-art method implemented on the DeeplabV2 architecture when combined with MIC.

\textbf{Synthia$\rightarrow$Cityscape.} We observed similar experimental results on the Synthia Benchmark, achieving scores of 55.6, 55.0, 62.8, and 63.7 on the DACS \cite{ref9}, DAFormer \cite{ref10}, HRDA \cite{ref11}, and MIC \cite{ref55} baselines, with respective improvements of +7.3, +0.8, +1.7, and +1.0. Additionally, besides aiding in distinguishing between motorcycles and bicycles, our approach effectively enhances segmentation in challenging categories like Rider and Person. For instance, on the HRDA architecture, we observed increases of +2.6 IOU and +5.0 IOU over the baseline for Rider and Person, respectively. Similarly, for Traffic Sign (Sign) and Traffic Light (Tr.Light), improvements of +3.2 IOU and +1.6 IOU, respectively, are noted. Overall, Guidance Training demonstrates consistent contributions to enhancing segmentation in challenging categories across most baselines, indicating its effectiveness in mitigating issues arising from cross-domain data mixing.

\subsection{Ablation Study}
\begin{table*}[ht]
	\caption{Ablation Study of Guidance Training. Guidance Training with DACS on GTA$\rightarrow$Cityscape Benchmark.}
	\begin{minipage}{0.3\linewidth}
		\vspace{-1.3 cm}
		\centering
		\renewcommand{\thetable}{IV.A}
		\setlength{\abovecaptionskip}{0.cm}
		\setlength{\belowcaptionskip}{-0.cm}
		\caption{Embedding Dim for Transformer Blocks.}
		\begin{tabular}{ccccc}
			\toprule
			Dim & 128 & 256 & 512 & 768\\
			\midrule
			mIOU & 55.5 & 55.0 & \textbf{55.9} & 54.9\\
			\bottomrule
		\end{tabular}
	\end{minipage}
	\hfill 
	\begin{minipage}{0.3\linewidth}
		\vspace{-1.3 cm}
		\setlength{\abovecaptionskip}{0.cm}
		\setlength{\belowcaptionskip}{-0.cm}
		\renewcommand{\thetable}{IV.B}
		\caption {Number of Transformer Blocks in Guider.}
		\begin{tabular}{cccccc}
			\toprule
			Blocks & 0 & 1 & 2 & 3 & 4 \\
			\midrule
			mIOU & 54.1 & 55.3 & \textbf{55.9} & 53.1 & 53.4 \\
			\bottomrule
		\end{tabular}
	\end{minipage}
	\hfill
	\begin{minipage}{0.3\linewidth}
		\vspace{-0.3 cm}
		\centering
		\setlength{\abovecaptionskip}{0.cm}
		\setlength{\belowcaptionskip}{-0.cm}
		\renewcommand{\thetable}{IV.C}
		\caption{The Role of Zero Convolution in Guider.}
		\begin{tabular}{cc|c}
			\toprule
			$Z_1(\cdot)$ & $Z_2(\cdot)$ & mIOU \\
			\midrule
			& &  53.7\\
			\ding{52} &  & 55.0 \\
			 & \ding{52} & 55.1  \\
			 \ding{52} & \ding{52} & \textbf{55.9}\\
			\bottomrule
		\end{tabular}
	\end{minipage}
	\hfill 
	\begin{minipage}{0.3\linewidth}
		\centering
		\setlength{\abovecaptionskip}{0.cm}
		\setlength{\belowcaptionskip}{-0.cm}
		\renewcommand{\thetable}{IV.D}
		\caption{Trade-off $\lambda_{gt}$ is Investigated.}
		\begin{tabular}{ccccc}
			\toprule
			$\lambda_{gt}$ & 0.1 & 0.5 & 1.0 & 2.0 \\
			\midrule
			mIOU &54.8 & 55.4 & \textbf{55.9} & 52.2 \\
			\bottomrule
		\end{tabular}
	\end{minipage}
	\hfill 
	\begin{minipage}{0.3\linewidth}
		\centering
		\setlength{\abovecaptionskip}{0.cm}
		\setlength{\belowcaptionskip}{-0.cm}
		\renewcommand{\thetable}{IV.E}
		\caption{$d$ in Uncertainty Estimation (UE).}
		\begin{tabular}{cccccc}
			\toprule
			$d$ & 0.0 & 2.0 & 5.0 & 10.0 & w/o UE\\
			\midrule
			mIOU &53.2 & 54.4 & \textbf{55.9} & 55.6 & 55.0 \\
			\bottomrule
		\end{tabular}
	\end{minipage}
	\hfill 
	\begin{minipage}{0.3\linewidth}
		\centering
		\setlength{\abovecaptionskip}{0.cm}
		\setlength{\belowcaptionskip}{-0.cm}
		\renewcommand{\thetable}{IV.F}
		\caption{The Role of Skip Connection in Guider.}
		\begin{tabular}{ccc}
			\toprule
			Skip Connection & \ding{55}  & \ding{52} \\
			\midrule
			mIOU &54.3 & \textbf{55.9}\\
			\bottomrule
		\end{tabular}
	\end{minipage}

\end{table*}
The design of Guidance Training will be thoroughly explored and analyzed in detail through ablation study, with detailed experimental results presented in Table IV. This analysis primarily encompasses hyper-parameter selection for Guider (Table IV.A–Table IV.C), the trade-off of Guidance Training (Table IV.D), the role of uncertainty estimation (Table IV.E), and the learning paradigms for pseudo target feature (Table IV.F). 

\textbf{Hyper-parameter selection for Guider.} In Section III.C, we delineate the architectural design of Guider and elucidate the rationale behind its design. The architecture of Guider is characterized by its simplicity, consisting of two zero-initialized convolutions and a GIA module. Within the GIA module, patch embedding and linear projection serve primarily for dimension reduction and expansion, respectively, to alleviate computational burden. As these operators are closely related to the embedding dim of Transformer blocks, we focus on studying the selection of Embedding Dim and do not analyze them separately. In Table IV.A, we investigate the Transformer across common four-dimensional settings and observe that there is little difference in accuracy among dimensions of 128, 256, 512 and 768. Consequently, we opt for an embedding dimension of 512. Next, we consider the choice of the number of Transformer Blocks, and we observe that the best results are achieved when the number of Blocks is set to 2. 

Interestingly, we also encounter a counterintuitive phenomenon: we anticipated that increasing the number of parameters in Guider (i.e., increasing the number of embedding dimensions and blocks) would lead to better performance. However, we find that when the number of blocks is set to 4, the performance decreases by -2.5 mIOU compared to $\text{Blocks}=2$. Similarly, when the embedding dimension is set to 768, the performance drops by -1.0 mIOU compared to $\text{Dim}=512$. We speculate that this phenomenon occurs because Guidance Training may induce overfitting when Guider's capacity is too high, leading to the model fitting noise in the pseudo-labels of the target image, consequently reducing overall performance.

In addition to the GIA module, we introduced two zero-initialization convolutions, $Z_1(\cdot)$ and $Z_2(\cdot)$, in Guider to stabilize training. In Table IV.C, we observe that utilizing either zero-initialization convolution yields significantly better results compared to not using any zero-initialization convolution. Moreover, optimal performance is achieved when both $Z_1(\cdot)$ and $Z_2(\cdot)$ are employed.

\textbf{The trade-off $\lambda_{gt}$ of Guidance Training.} Additionally, we investigate the weights $\lambda_{gt}$ of Guidance Training losses $\mathcal{L}_{gt}$. In Table IV.D, the optimal performance is attained when the weights is set to 1.0. However, as the weights gradually decrease, we observe a corresponding gradual decline in performance. Notably, when $\lambda_{gt}$ is set to 2.0, there is a significant decrease in performance. We hypothesize that this decline in performance, akin to that caused by the intricate design of Guider, is attributed to the model overfitting to the pseudo-label learning process.

\textbf{The effect of uncertainty estimation.} In Section III.B, we introduce the main idea of Guidance Training, which predicts the pseudo-labeling of the target domain image by introducing the Guider to transform the hybrid feature $E(x^{m})$ and decoding $G(E(x^{m}),M)$ by decoder. Since the ratio of the source domain to the target domain is random (controlled by DACS \cite{ref9}) in a hybrid image, it is more difficult to predict the target domain pseudo-labels when the ratio of the source domain is larger. Therefore, in Eq. (12), we introduce uncertainty estimation, and we expect the dynamic factor $\beta$ to adjust the weights dynamically when the proportion of the source domain is larger. In Eq. (12), $\beta$ is mainly affected by the parameter $d$, and when $d$ is larger, $\beta$ is less sensitive to uncertainty. In Table IV.E, we find that the best performance is achieved when $d$ is chosen to be 5.0. When no uncertainty estimation is used, the performance drops by -0.9 mIOU.

\textbf{Learning paradigms for pseudo target feature.} In Section III.C, we explore the design principles of Guider in detail, and ultimately, we opt to model pseudo-target features in a learning-offset manner. In Table IV.F, we compare the direct learning of pseudo-target features (\ding{55} Skip Connection) with the learning-offset approach (\ding{52} Skip Connection). When the learning-offset approach is employed, it improves +1.6 mIOU over the direct learning of features.

\subsection{Qualitative Analysis}
\begin{figure*}[!t]
	\setlength{\abovecaptionskip}{0cm}  
	\setlength{\belowcaptionskip}{-0.2cm} 
	\centering
	\includegraphics[width=7.0in]{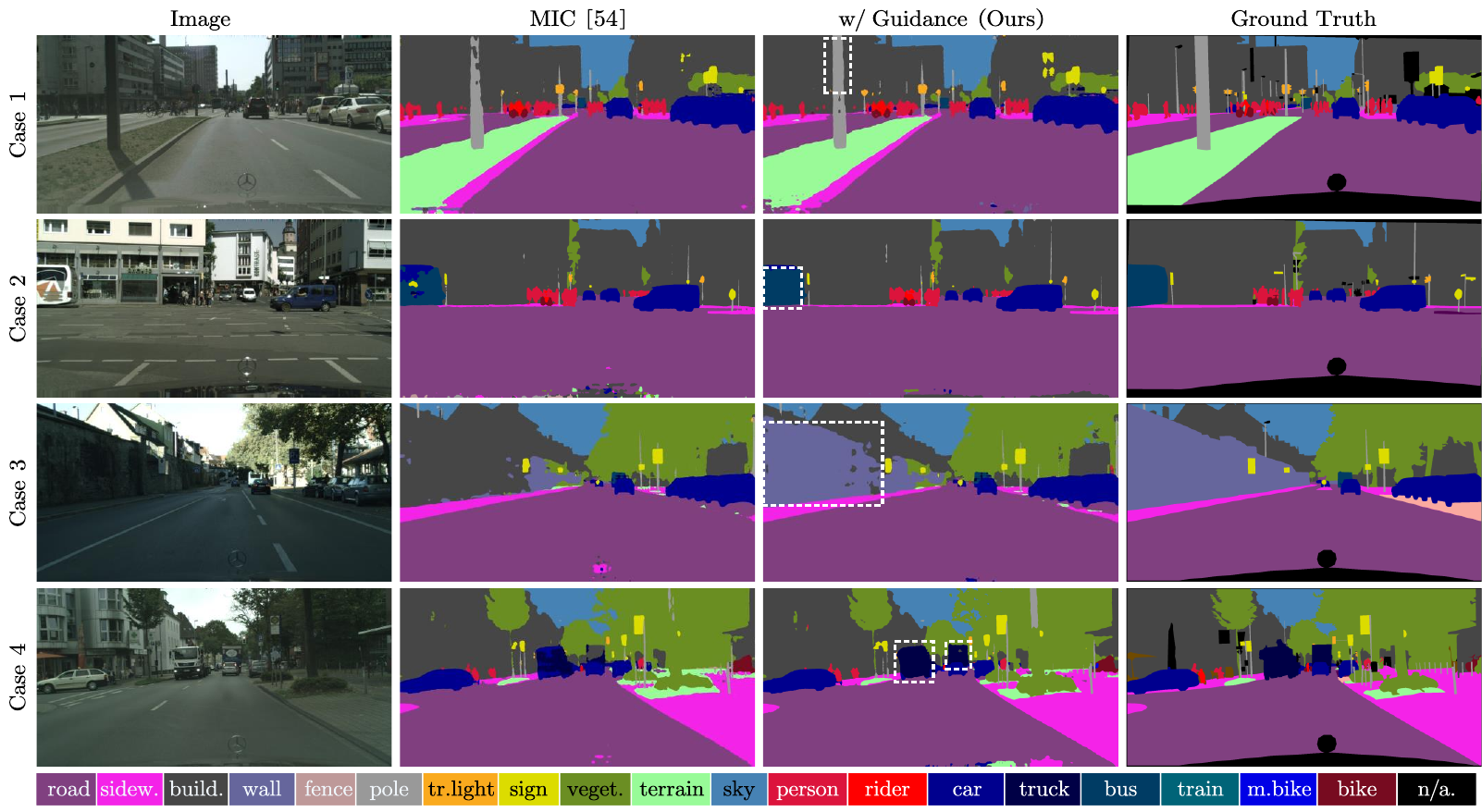}
	\caption{Qualitative comparison of Guidance Training with MIC\cite{ref55} on GTA$\rightarrow$Cityscape. We combine Guidance Training with MIC \cite{ref55} and find, through visualization, that Guidance Training prevents the model from deviating from the physical world distribution. For details of the analysis, refer to Section IV.E.}
\end{figure*}
\begin{figure*}[!t]
	\setlength{\abovecaptionskip}{0cm}  
	\setlength{\belowcaptionskip}{-0.2cm} 
	\centering
	\includegraphics[width=7.0in]{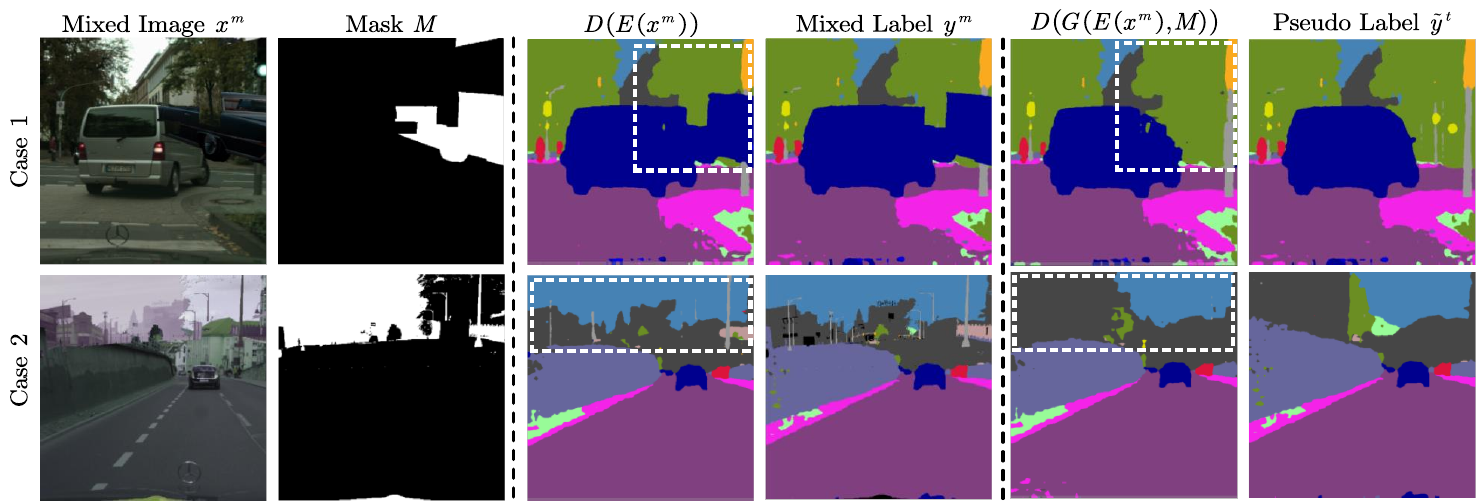}
	\caption{Visual examples of mixed data prediction  $D(E(x^m))$ and pseudo-label prediction based on mixed feature $D(G(E(x^m),M))$.}
\end{figure*}

In this section, we primarily aim to analyze two questions. The first question pertains to how Guidance Training enhances the cross-domain data mixing based algorithms. The second question examines whether Guidance Training truly endows the model with the capability to predict the pseudo-labels of the target image from the mixed image, while accurately predicting the mixed labels of the mixed image simultaneously. If the model successfully accomplishes the task posed in the second question, it signifies two key aspects. Firstly, it demonstrates that the model comprehensively understands the real-world physical distribution, as it can decouple the physical world distribution from the mixed data. Secondly, it confirms that the model can still effectively utilize mixed data to overcome the domain gap.

\textbf{Question1: How Guidance Training enhances the cross-domain data mixing based algorithms?} We combine MIC \cite{ref55} with Guidance Training and display the prediction results for four examples in Figure 3.

In case 1, MIC's prediction of the pole shows a discontinuity with many holes, a phenomenon unimaginable in the physical world. When combined with our method, the prediction of the pole exhibits continuity.

In case 2, MIC predicts the body of a bus as a traffic sign, whereas our method successfully rectifies this result, aligning with common sense.

In case 3, MIC predicts a large area as a building instead of a continuous wall, contrary to human visual perception. Guidance Training effectively captures contextual properties of visual semantics and corrects MIC's misprediction.

In case 4, MIC's prediction result for two trucks violates physical common sense. While half of the trucks on the left are predicted as cars, even half of the trucks on the right are predicted to be buildings. This inconsistency is rectified when combined with Guidance Training, leading to correct predictions.

In summary, Figure 3 illustrates instances where MIC makes predictions that defy common sense in the physical world. This can partly be attributed to the mixed data distribution introduced by data-mixing techniques, which disrupts the real-world distribution and diminishes model robustness. Through our method, we enhance the cross-domain data mixing technique, thereby improving model robustness.

\textbf{Question2: Can Guidance Training and mixed data prediction collaborate effectively?} To verify this, we select two cases where mixed data is input into the model to obtain the prediction result $D(E(x^m))$. Concurrently, we introduce Guider to derive the pseudo-labeled prediction result $D(G(E(x^m),M))$ based on mixed features.

In case 1, we observe a car floating in mid-air in the mixed data. $D(E(x^m))$ effectively predict the mixed label. Meanwhile, in $D(G(E(x^m),M))$, the model successfully remove the floating car and accurately predict the vegetation and pole occluded by the car.

In case 2, the fence and sky are pasted onto the image. $D(E(x^m))$ successfully predict the image details within the white dashed box, including regions of the pole, fence, and sky. In $D(G(E(x^m),M))$, the model accurately predict the complete building.

Overall, we observe the following phenomena:

Firstly, from the blended image $x^m$, we note numerous regions contradicting the real-world distribution. Direct fitting of these regions would lead to significant deviation from the real-world distribution. Secondly, even after the introduction of Guidance Training, the model remain capable of effectively utilizing blended data to overcome the domain gap. Finally, the pseudo-labeling prediction $D(G(E(x^m),M))$ based on hybrid features accurately predict pseudo-label. This suggests that the model's representations possess the ability to extract real-world norms from hybrid data distributions and reconstruct real-data distributions, thereby avoiding distributional bias in the model.

\section{Conclusion and Future Works}
In this paper, we introduce Guidance Training as a method to improve the Cross-domain mixed sampling, thereby boosting the performance and robustness of models when integrated with popular methods. Our primary contribution lies in identifying the limitations of existing popular approaches baseline and addressing the potential distributional bias introduced by mix sampling. Importantly, our method achieves these improvements with minimal training overhead and without increasing the inference burden.

While we demonstrate the simplicity and effectiveness of Guidance Training in this paper, we recognize significant potential for future exploration. Firstly, we aim to investigate whether Guidance Training can be applied to more general architectures, such as non-encoder-decoder structures, multi-scale models. Secondly, considering that mixed sampling is utilized not only in domain adaptive segmentation but also in various other domains, we plan to explore the applicability of Guidance Training in these contexts. Finally, as the core objective of Guidance Training is to extract and reconstruct target features from mixed features, we intend to explore additional methods to further enhance performance efficiency.

Moving forward, we are committed to exploring and addressing these questions to advance the field of domain-adaptive segmentation and improve model robustness.

\bibliographystyle{IEEEtran}
\bibliography{reference}

\begin{thebibliography}{10}
\providecommand{\url}[1]{#1}
\csname url@samestyle\endcsname
\providecommand{\newblock}{\relax}
\providecommand{\bibinfo}[2]{#2}
\providecommand{\BIBentrySTDinterwordspacing}{\spaceskip=0pt\relax}
\providecommand{\BIBentryALTinterwordstretchfactor}{4}
\providecommand{\BIBentryALTinterwordspacing}{\spaceskip=\fontdimen2\font plus
\BIBentryALTinterwordstretchfactor\fontdimen3\font minus
  \fontdimen4\font\relax}
\providecommand{\BIBforeignlanguage}[2]{{%
\expandafter\ifx\csname l@#1\endcsname\relax
\typeout{** WARNING: IEEEtran.bst: No hyphenation pattern has been}%
\typeout{** loaded for the language `#1'. Using the pattern for}%
\typeout{** the default language instead.}%
\else
\language=\csname l@#1\endcsname
\fi
#2}}
\providecommand{\BIBdecl}{\relax}
\BIBdecl

\bibitem{ref1}
B.~Cheng, W.~Wu, D.~Tao, S.~Mei, T.~Mao, and J.~Cheng, ``Random cropping
  ensemble neural network for image classification in a robotic arm grasping
  system,'' \emph{IEEE Trans. Instrum. Meas.}, vol.~69, no.~9, pp. 6795--6806,
  2020.

\bibitem{ref2}
J.~Luo, Z.~Yang, S.~Li, and Y.~Wu, ``Fpcb surface defect detection: A decoupled
  two-stage object detection framework,'' \emph{IEEE Trans. Instrum. Meas.},
  vol.~70, pp. 1--11, 2021.

\bibitem{ref3}
W.~Xie, P.~X. Liu, and M.~Zheng, ``Moving object segmentation and detection for
  robust rgbd-slam in dynamic environments,'' \emph{IEEE Trans. Instrum.
  Meas.}, vol.~70, pp. 1--8, 2020.

\bibitem{ref4}
G.~Wilson and D.~J. Cook, ``A survey of unsupervised deep domain adaptation,''
  \emph{ACM Trans. Intell. Syst. Technol.}, vol.~11, no.~5, pp. 1--46, 2020.

\bibitem{ref5}
M.~Toldo, A.~Maracani, U.~Michieli, and P.~Zanuttigh, ``Unsupervised domain
  adaptation in semantic segmentation: a review,'' \emph{Technologies}, vol.~8,
  no.~2, p.~35, 2020.

\bibitem{ref6}
S.~Yun, D.~Han, S.~J. Oh, S.~Chun, J.~Choe, and Y.~Yoo, ``Cutmix:
  Regularization strategy to train strong classifiers with localizable
  features,'' in \emph{IEEE Int. Conf. Comput. Vis. (ICCV)}, 2019, pp.
  6023--6032.

\bibitem{ref7}
H.~Zhang, M.~Cisse, Y.~N. Dauphin, and D.~Lopez-Paz, ``mixup: Beyond empirical
  risk minimization,'' in \emph{Int. Conf. Learn. Represent. (ICLR)}, 2017.

\bibitem{ref8}
D.~Berthelot, N.~Carlini, I.~Goodfellow, N.~Papernot, A.~Oliver, and C.~A.
  Raffel, ``Mixmatch: A holistic approach to semi-supervised learning,'' in
  \emph{Adv. Neural Inf. Process. Syst. (NIPS)}, 2019.

\bibitem{ref9}
W.~Tranheden, V.~Olsson, J.~Pinto, and L.~Svensson, ``Dacs: Domain adaptation
  via cross-domain mixed sampling,'' in \emph{WACV}, 2021, pp. 1379--1389.

\bibitem{ref10}
L.~Hoyer, D.~Dai, and L.~Van~Gool, ``Daformer: Improving network architectures
  and training strategies for domain-adaptive semantic segmentation,'' in
  \emph{IEEE Conf. Comput. Vis. Pattern Recognit. (CVPR)}, 2022, pp.
  9924--9935.

\bibitem{ref11}
{L. Hoyer, D. Dai and L. Van Gool}, ``Hrda: Context-aware high-resolution
  domain-adaptive semantic segmentation,'' in \emph{Eur. Conf. Comput. Vis.
  (ECCV)}.\hskip 1em plus 0.5em minus 0.4em\relax Springer, 2022, pp. 372--391.

\bibitem{ref12}
X.~Huo, L.~Xie, H.~Hu, W.~Zhou, H.~Li, and Q.~Tian, ``Domain-agnostic prior for
  transfer semantic segmentation,'' in \emph{IEEE Conf. Comput. Vis. Pattern
  Recognit. (CVPR)}, 2022, pp. 7075--7085.

\bibitem{ref13}
S.~Saha, L.~Hoyer, A.~Obukhov, D.~Dai, and L.~Van~Gool, ``Edaps: Enhanced
  domain-adaptive panoptic segmentation,'' in \emph{IEEE Int. Conf. Comput.
  Vis. (ICCV)}, 2023.

\bibitem{ref14}
R.~Xia, C.~Zhao, M.~Zheng, Z.~Wu, Q.~Sun, and Y.~Tang, ``Cmda: Cross-modality
  domain adaptation for nighttime semantic segmentation,'' in \emph{IEEE Int.
  Conf. Comput. Vis. (ICCV)}, 2023, pp. 21\,572--21\,581.

\bibitem{ref18}
J.~Long, E.~Shelhamer, and T.~Darrell, ``Fully convolutional networks for
  semantic segmentation,'' in \emph{IEEE Conf. Comput. Vis. Pattern Recognit.
  (CVPR)}, 2015, pp. 3431--3440.

\bibitem{ref19}
O.~Ronneberger, P.~Fischer, and T.~Brox, ``U-net: Convolutional networks for
  biomedical image segmentation,'' in \emph{MICCAI}, 2015, pp. 234--241.

\bibitem{ref20}
L.-C. Chen, G.~Papandreou, I.~Kokkinos, K.~Murphy, and A.~L. Yuille, ``Semantic
  image segmentation with deep convolutional nets and fully connected crfs,''
  \emph{arXiv preprint arXiv:1412.7062}, 2014.

\bibitem{ref21}
{L. Chen, G. Papandreou, I. Kokkinos, K. Murphy and A. Yuille}, ``Deeplab:
  Semantic image segmentation with deep convolutional nets, atrous convolution,
  and fully connected crfs,'' \emph{IEEE Trans. Pattern Anal. Mach. Intell.},
  vol.~40, no.~4, pp. 834--848, 2017.

\bibitem{ref22}
L.-C. Chen, Y.~Zhu, G.~Papandreou, F.~Schroff, and H.~Adam, ``Encoder-decoder
  with atrous separable convolution for semantic image segmentation,'' in
  \emph{Eur. Conf. Comput. Vis. (ECCV)}, 2018, pp. 801--818.

\bibitem{ref23}
Y.~LeCun, L.~Bottou, Y.~Bengio, and P.~Haffner, ``Gradient-based learning
  applied to document recognition,'' \emph{Proc. of the IEEE}, vol.~86, no.~11,
  pp. 2278--2324, 1998.

\bibitem{ref24}
A.~Vaswani, N.~Shazeer, N.~Parmar, J.~Uszkoreit, L.~Jones, A.~N. Gomez,
  {\L}.~Kaiser, and I.~Polosukhin, ``Attention is all you need,'' in \emph{Adv.
  Neural Inf. Process. Syst. (NIPS)}, 2017.

\bibitem{ref25}
X.~Li, H.~Ding, W.~Zhang, H.~Yuan, J.~Pang, G.~Cheng, K.~Chen, Z.~Liu, and
  C.~C. Loy, ``Transformer-based visual segmentation: A survey,'' \emph{arXiv
  preprint arXiv:2304.09854}, 2023.

\bibitem{ref26}
E.~Xie, W.~Wang, Z.~Yu, A.~Anandkumar, J.~M. Alvarez, and P.~Luo, ``Segformer:
  Simple and efficient design for semantic segmentation with transformers,''
  vol.~34, 2021, pp. 12\,077--12\,090.

\bibitem{ref27}
Z.~Liu, H.~Mao, C.-Y. Wu, C.~Feichtenhofer, T.~Darrell, and S.~Xie, ``A convnet
  for the 2020s,'' in \emph{IEEE Conf. Comput. Vis. Pattern Recognit. (CVPR)},
  2022, pp. 11\,976--11\,986.

\bibitem{ref28}
M.-H. Guo, C.-Z. Lu, Q.~Hou, Z.~Liu, M.-M. Cheng, and S.-M. Hu, ``Segnext:
  Rethinking convolutional attention design for semantic segmentation,''
  \emph{Adv. Neural Inf. Process. Syst. (NIPS)}, vol.~35, pp. 1140--1156, 2022.

\bibitem{ref29}
I.~Goodfellow, J.~Pouget-Abadie, M.~Mirza, B.~Xu, D.~Warde-Farley, S.~Ozair,
  A.~Courville, and Y.~Bengio, ``Generative adversarial nets,'' in \emph{Adv.
  Neural Inf. Process. Syst. (NIPS)}, 2014.

\bibitem{ref30}
H.~Rangwani, S.~K. Aithal, M.~Mishra, A.~Jain, and V.~B. Radhakrishnan, ``A
  closer look at smoothness in domain adversarial training,'' in \emph{Proc.
  Int. Conf. Mach. Learn. (ICML)}, 2022, pp. 18\,378--18\,399.

\bibitem{ref31}
H.~Liu, J.~Wang, and M.~Long, ``Cycle self-training for domain adaptation,'' in
  \emph{Proc. Adv. Neural Inf. Process. Syst. (NIPS)}, 2021, pp.
  22\,968--22\,981.

\bibitem{ref32}
Z.~Deng, Y.~Luo, and J.~Zhu, ``Cluster alignment with a teacher for
  unsupervised domain adaptation,'' in \emph{Proc. IEEE Int. Conf. Comput. Vis.
  (ICCV)}, 2019, pp. 9944--9953.

\bibitem{ref33}
M.~Chen, Z.~Zheng, Y.~Yang, and T.-S. Chua, ``Pipa: Pixel-and patch-wise
  self-supervised learning for domain adaptative semantic segmentation,'' in
  \emph{ACM MM}, 2023, pp. 1905--1914.

\bibitem{ref47}
H.~Guo, Y.~Mao, and R.~Zhang, ``Augmenting data with mixup for sentence
  classification: An empirical study,'' \emph{arXiv preprint arXiv:1905.08941},
  2019.

\bibitem{ref48}
L.~Sun, C.~Xia, W.~Yin, T.~Liang, P.~S. Yu, and L.~He, ``Mixup-transformer:
  dynamic data augmentation for nlp tasks,'' \emph{arXiv preprint
  arXiv:2010.02394}, 2020.

\bibitem{ref49}
V.~Verma, A.~Lamb, C.~Beckham, A.~Najafi, I.~Mitliagkas, D.~Lopez-Paz, and
  Y.~Bengio, ``Manifold mixup: Better representations by interpolating hidden
  states,'' in \emph{Proc. Int. Conf. Mach. Learn. (ICML)}.\hskip 1em plus
  0.5em minus 0.4em\relax PMLR, 2019, pp. 6438--6447.

\bibitem{ref40}
V.~Olsson, W.~Tranheden, J.~Pinto, and L.~Svensson, ``Classmix:
  Segmentation-based data augmentation for semi-supervised learning,'' in
  \emph{WACV}, 2021, pp. 1369--1378.

\bibitem{ref34}
J.~Devlin, M.-W. Chang, K.~Lee, and K.~Toutanova, ``Bert: Pre-training of deep
  bidirectional transformers for language understanding,'' 2018.

\bibitem{ref35}
Y.~Liu, M.~Ott, N.~Goyal, J.~Du, M.~Joshi, D.~Chen, O.~Levy, M.~Lewis,
  L.~Zettlemoyer, and V.~Stoyanov, ``Roberta: A robustly optimized bert
  pretraining approach,'' \emph{arXiv preprint arXiv:1907.11692}, 2019.

\bibitem{ref17}
K.~He, X.~Chen, S.~Xie, Y.~Li, P.~Doll{\'a}r, and R.~Girshick, ``Masked
  autoencoders are scalable vision learners,'' in \emph{IEEE Conf. Comput. Vis.
  Pattern Recognit. (CVPR)}, 2022, pp. 16\,000--16\,009.

\bibitem{ref36}
C.~Wei, H.~Fan, S.~Xie, C.-Y. Wu, A.~Yuille, and C.~Feichtenhofer, ``Masked
  feature prediction for self-supervised visual pre-training,'' in \emph{IEEE
  Conf. Comput. Vis. Pattern Recognit. (CVPR)}, 2022, pp. 14\,668--14\,678.

\bibitem{ref37}
S.~Woo, S.~Debnath, R.~Hu, X.~Chen, Z.~Liu, I.~S. Kweon, and S.~Xie, ``Convnext
  v2: Co-designing and scaling convnets with masked autoencoders,'' in
  \emph{IEEE Conf. Comput. Vis. Pattern Recognit. (CVPR)}, 2023, pp.
  16\,133--16\,142.

\bibitem{ref15}
H.~Bao, L.~Dong, S.~Piao, and F.~Wei, ``Beit: Bert pre-training of image
  transformers,'' in \emph{Int. Conf. Learn. Represent. (ICLR)}, 2021.

\bibitem{ref38}
A.~Van Den~Oord, O.~Vinyals \emph{et~al.}, ``Neural discrete representation
  learning,'' in \emph{Adv. Neural Inf. Process. Syst. (NIPS)}, vol.~30, 2017.

\bibitem{ref39}
M.~Assran, Q.~Duval, I.~Misra, P.~Bojanowski, P.~Vincent, M.~Rabbat, Y.~LeCun,
  and N.~Ballas, ``Self-supervised learning from images with a joint-embedding
  predictive architecture,'' in \emph{IEEE Conf. Comput. Vis. Pattern Recognit.
  (CVPR)}, 2023, pp. 15\,619--15\,629.

\bibitem{ref41}
A.~Tarvainen and H.~Valpola, ``Mean teachers are better role models:
  Weight-averaged consistency targets improve semi-supervised deep learning
  results,'' in \emph{Adv. Neural Inf. Process. Syst. (NIPS)}, 2017.

\bibitem{ref42}
M.~Ghifary, W.~B. Kleijn, M.~Zhang, D.~Balduzzi, and W.~Li, ``Deep
  reconstruction-classification networks for unsupervised domain adaptation,''
  in \emph{Eur. Conf. Comput. Vis. (ECCV)}.\hskip 1em plus 0.5em minus
  0.4em\relax Springer, 2016, pp. 597--613.

\bibitem{ref43}
M.~Long, Y.~Cao, J.~Wang, and M.~Jordan, ``Learning transferable features with
  deep adaptation networks,'' in \emph{Proc. Int. Conf. Mach. Learn. (ICML)},
  2015, pp. 97--105.

\bibitem{ref44}
Y.~Ganin and V.~Lempitsky, ``Unsupervised domain adaptation by
  backpropagation,'' in \emph{Proc. Int. Conf. Mach. Learn. (ICML)}, 2015, pp.
  1180--1189.

\bibitem{ref45}
A.~Dosovitskiy, L.~Beyer, A.~Kolesnikov, D.~Weissenborn, X.~Zhai,
  T.~Unterthiner, M.~Dehghani, M.~Minderer, G.~Heigold, S.~Gelly \emph{et~al.},
  ``An image is worth 16x16 words: Transformers for image recognition at
  scale,'' in \emph{Proc. Int. Conf. Learn. Represent. (ICLR)}, 2021.

\bibitem{ref46}
L.~Zhang, A.~Rao, and M.~Agrawala, ``Adding conditional control to
  text-to-image diffusion models,'' in \emph{IEEE Int. Conf. Comput. Vis.
  (ICCV)}, 2023, pp. 3836--3847.

\bibitem{ref50}
S.~R. Richter, V.~Vineet, S.~Roth, and V.~Koltun, ``Playing for data: Ground
  truth from computer games,'' in \emph{Eur. Conf. Comput. Vis. (ECCV)}.\hskip
  1em plus 0.5em minus 0.4em\relax Springer, 2016, pp. 102--118.

\bibitem{ref51}
G.~Ros, L.~Sellart, J.~Materzynska, D.~Vazquez, and A.~M. Lopez, ``The synthia
  dataset: A large collection of synthetic images for semantic segmentation of
  urban scenes,'' in \emph{IEEE Conf. Comput. Vis. Pattern Recognit. (CVPR)},
  2016, pp. 3234--3243.

\bibitem{ref52}
M.~Cordts, M.~Omran, S.~Ramos, T.~Rehfeld, M.~Enzweiler, R.~Benenson,
  U.~Franke, S.~Roth, and B.~Schiele, ``The cityscapes dataset for semantic
  urban scene understanding,'' in \emph{IEEE Conf. Comput. Vis. Pattern
  Recognit. (CVPR)}, 2016, pp. 3213--3223.

\bibitem{ref53}
K.~He, X.~Zhang, S.~Ren, and J.~Sun, ``Deep residual learning for image
  recognition,'' in \emph{IEEE Conf. Comput. Vis. Pattern Recognit. (CVPR)},
  2016, pp. 770--778.

\bibitem{ref54}
I.~Loshchilov and F.~Hutter, ``Decoupled weight decay regularization,'' in
  \emph{Proc. Int. Conf. Learn. Represent. (ICLR)}, 2018.

\bibitem{ref55}
L.~Hoyer, D.~Dai, H.~Wang, and L.~Van~Gool, ``Mic: Masked image consistency for
  context-enhanced domain adaptation,'' in \emph{IEEE Conf. Comput. Vis.
  Pattern Recognit. (CVPR)}, 2023, pp. 11\,721--11\,732.

\bibitem{ref56}
K.~Mei, C.~Zhu, J.~Zou, and S.~Zhang, ``Instance adaptive self-training for
  unsupervised domain adaptation,'' in \emph{Eur. Conf. Comput. Vis.
  (ECCV)}.\hskip 1em plus 0.5em minus 0.4em\relax Springer, 2020, pp. 415--430.

\bibitem{ref57}
Y.~Wang, J.~Peng, and Z.~Zhang, ``Uncertainty-aware pseudo label refinery for
  domain adaptive semantic segmentation,'' in \emph{IEEE Int. Conf. Comput.
  Vis. (ICCV)}, 2021, pp. 9092--9101.

\bibitem{ref58}
Y.~Cheng, F.~Wei, J.~Bao, D.~Chen, F.~Wen, and W.~Zhang, ``Dual path learning
  for domain adaptation of semantic segmentation,'' in \emph{IEEE Int. Conf.
  Comput. Vis. (ICCV)}, 2021, pp. 9082--9091.

\bibitem{ref59}
N.~Araslanov and S.~Roth, ``Self-supervised augmentation consistency for
  adapting semantic segmentation,'' in \emph{IEEE Conf. Comput. Vis. Pattern
  Recognit. (CVPR)}, 2021, pp. 15\,384--15\,394.

\bibitem{ref60}
H.~Ma, X.~Lin, Z.~Wu, and Y.~Yu, ``Coarse-to-fine domain adaptive semantic
  segmentation with photometric alignment and category-center regularization,''
  in \emph{IEEE Conf. Comput. Vis. Pattern Recognit. (CVPR)}, 2021, pp.
  4051--4060.

\bibitem{ref61}
Q.~Wang, D.~Dai, L.~Hoyer, L.~Van~Gool, and O.~Fink, ``Domain adaptive semantic
  segmentation with self-supervised depth estimation,'' in \emph{IEEE Int.
  Conf. Comput. Vis. (ICCV)}, 2021, pp. 8515--8525.

\bibitem{ref62}
P.~Zhang, B.~Zhang, T.~Zhang, D.~Chen, Y.~Wang, and F.~Wen, ``Prototypical
  pseudo label denoising and target structure learning for domain adaptive
  semantic segmentation,'' in \emph{IEEE Conf. Comput. Vis. Pattern Recognit.
  (CVPR)}, 2021, pp. 12\,414--12\,424.

\bibitem{ref63}
B.~Xie, S.~Li, M.~Li, C.~H. Liu, G.~Huang, and G.~Wang, ``Sepico:
  Semantic-guided pixel contrast for domain adaptive semantic segmentation,''
  \emph{IEEE Trans. Pattern Anal. Mach. Intell.}, 2023.

\bibitem{ref64}
J.~Lu, J.~Shi, H.~Zhu, J.~Ni, X.~Shu, Y.~Sun, and Z.~Cheng, ``Depth guidance
  and intradomain adaptation for semantic segmentation,'' \emph{IEEE Trans.
  Instrum. Meas.}, vol.~72, pp. 1--13, 2023.

\end{thebibliography}
	
\vfill

\end{document}